\documentclass[10pt,twocolumn,letterpaper]{article}

\usepackage[pagenumbers]{wacv} 

\usepackage[table,x11names,dvipsnames,svgnames]{xcolor}
\usepackage{colortbl}
\usepackage{graphicx}
\usepackage{amsmath}
\usepackage{amssymb}
\usepackage{booktabs}
\usepackage[lined,boxed,commentsnumbered,ruled,linesnumbered,vlined]{algorithm2e}
\usepackage{multirow}
\usepackage{array}
\newcolumntype{P}[1]{>{\centering\arraybackslash}p{#1}}
\usepackage{tikz}
\usepackage{algorithmic}
\usepackage{tabularx}
\usepackage{pifont}

\usepackage[accsupp]{axessibility}

\newcommand{\cmark}{\ding{51}}
\newcommand{\xmark}{\ding{55}}
\definecolor{cmark}{RGB}{50,210,60}
\definecolor{xmark}{RGB}{210,50,60}
\newcommand{\cmarkcol}{\textcolor{cmark}{\cmark}}
\newcommand{\xmarkcol}{\textcolor{xmark}{\xmark}}
\newcommand{\tri}{\begin{small}\color{Green}$\blacktriangle$\end{small}}
\newcommand{\trid}{\begin{small}\color{BrickRed}$\blacktriangledown$\end{small}}

\newcommand\blfootnote[1]{%
  \begingroup
  \renewcommand\thefootnote{}\footnote{#1}%
  \addtocounter{footnote}{-1}%
  \endgroup
}

\usepackage[pagebackref,breaklinks,colorlinks]{hyperref}

\usepackage[capitalize]{cleveref}
\crefname{section}{Sec.}{Secs.}
\Crefname{section}{Section}{Sections}
\Crefname{table}{Table}{Tables}
\crefname{table}{Tab.}{Tabs.}

\def\wacvPaperID{1268} 
\def\confName{WACV}
\def\confYear{2025}

\begin{document}

\title{Self-Supervised Learning with Probabilistic Density Labeling \\for Rainfall Probability Estimation}

\author{
Junha Lee$^{1,3*}$, Sojung An$^{2*}$, Sujeong You$^{3}$, \vspace{0.051in} Namik Cho$^{1}$ \\
$^{1}$Seoul National University \\
$^{2}$Korea Institute of Atmospheric Prediction Systems \\
$^{3}$Korea Institute of Industrial Technology \\
{\tt\small \{junhalee,nicho\}@snu.ac.kr, sojungan@kiaps.org, sjyou21@kitech.re.kr}
}

\maketitle

\begin{abstract}
    \blfootnote{* Equal contribution}
    Numerical weather prediction (NWP) models are fundamental in meteorology for simulating and forecasting the behavior of various atmospheric variables. 
    The accuracy of precipitation forecasts and the acquisition of sufficient lead time are crucial for preventing hazardous weather events. 
    However, the performance of NWP models is limited by the nonlinear and unpredictable patterns of extreme weather phenomena driven by temporal dynamics.
    In this regard, we propose a \textbf{S}elf-\textbf{S}upervised \textbf{L}earning with \textbf{P}robabilistic \textbf{D}ensity \textbf{L}abeling (SSLPDL) for estimating rainfall probability by post-processing NWP forecasts.    
    Our post-processing method uses self-supervised learning (SSL) with masked modeling for reconstructing atmospheric physics variables, enabling the model to learn the dependency between variables. 
    The pre-trained encoder is then utilized in transfer learning to a precipitation segmentation task. 
    Furthermore, we introduce a straightforward labeling approach based on probability density to address the class imbalance in extreme weather phenomena like heavy rain events. 
    Experimental results show that SSLPDL surpasses other precipitation forecasting models in regional precipitation post-processing and demonstrates competitive performance in extending forecast lead times.
    Our code is available at \url{https://github.com/joonha425/SSLPDL}.
\end{abstract}

\begin{figure}[t]
    \centering
    \includegraphics[width=\columnwidth]{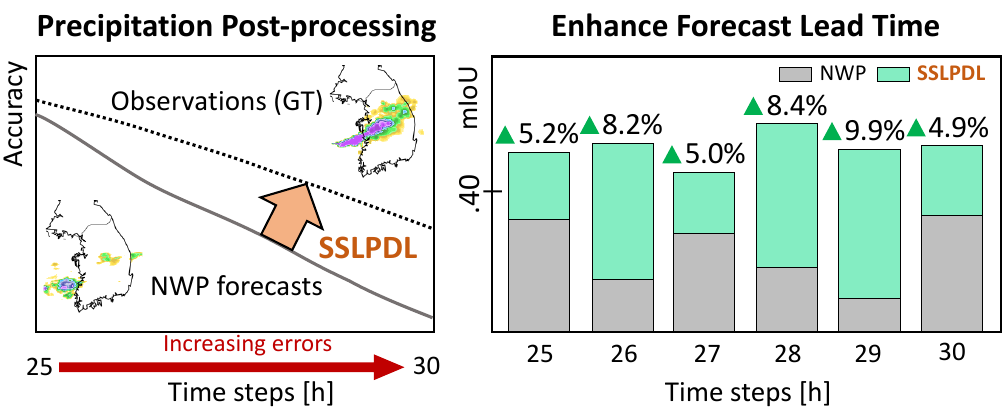}
    \caption{Our SSLPDL improves NWP forecast accuracy and secures extended lead times. \textbf{NWP forecasts refer to the predictions of future weather variables (e.g., temperature, humidity, rain) generated by NWP models.} The precipitation dataset from the NWP models was extracted for post-processing with a 24-hour lead time, using forecasts from 25 to 30 hours. Our approach consistently improves corrected rainfall across all evaluated lead times, showcasing its robustness and reliability in enhancing forecast accuracy.}
    \label{fig:problem}
\end{figure}

\section{Introduction}
\label{sec:intro}

NWP models forecast future atmospheric variables (e.g., temperature, wind speed, and precipitation) by solving fundamental equations of dynamics and physics.
These models discretize the atmosphere into 3D grid cubes and apply numerical methods to predict future atmospheric conditions over time.
Due to climate change, NWP models have become crucial for meteorologists worldwide to predict weather patterns and prevent economic damage~\cite{zheng2014urban,bendre2015big,smith2013environmental}.
The atmosphere is a chaotic system in which minor perturbations in initial conditions can lead to unpredictable outcomes over time.
More accurate predictions from NWP models depend on increasing the resolution of the input data~\cite{rojas2023deep}.
However, improving resolution is limited by computational costs and data processing constraints~\cite{ben2024rise,serifi2021spatio}, and accurately representing small-scale meteorological features such as convective phenomena remains challenging~\cite{de2023machine}.
Given these constraints, accumulated errors in NWP model precipitation predictions make securing sufficient lead time challenging.
To address these issues, it highlights the need for effective post-processing of NWP forecasts.

Recent studies have demonstrated the effectiveness of machine learning (ML) in reducing spatiotemporal biases in forecasts by capturing variable dependencies~\cite{kim:2022,rojas:2023,liu:2023}.
Consequently, ML-based precipitation post-processing has emerged as an active research area, aiming to enhance prediction performance over time by learning the correlations among atmospheric variables.
Several ML models widely used in computer vision have been proposed to boost the precipitation post-processing, such as artificial neural networks \cite{rojas:2023}, convolutional neural networks \cite{kim:2022, liu:2022, hess:2022}, and Transformers \cite{tang:2023}.
However, prior studies have focused on modeling simple state transitions of atmospheric variables at spatial locations. 
This approach constrains the ability to capture the dynamic spatiotemporal biases inherent in each variable and restricts the analysis of the non-stationarity present in diverse real-world data distributions.
Additionally, addressing class imbalance issues in precipitation datasets presents a critical challenge.
When some classes are sparsely distributed compared to others, the model tends to overfit the majority class during training, resulting in poor performance for the minority class. 
Extreme weather phenomena such as heavy rain are rare~\cite{davenport:2021,liu:2022real}, leading to class imbalance issues.

In this paper, we propose a \textbf{S}elf-\textbf{S}upervised \textbf{L}earning with \textbf{P}robabilistic \textbf{D}ensity \textbf{L}abeling (SSLPDL) for rainfall probability estimation by post-processing NWP forecasts.
We found that aggregating features from neighboring pixels better estimates the spatiotemporal bias in forecasts than learning from fixed pixels.
Our framework employs a pre-training process that can learn variable dependency using deformable convolution layers.
This pre-training framework flexibly aggregates information from neighboring pixels, learns bias from distribution shifts among variables, and addresses class imbalance issues through probabilistic density labeling.
We summarize the contributions of this work as below:
\begin{itemize}
    \item We design a self-supervised pre-training process that considers the physical and dynamic processes to correct the non-stationarity of spurious correlations among variables and reduce precipitation bias.
    \item We propose a probabilistic density labeling approach to address class imbalance. 
    The proposed labeling enhances generalization by preventing the overfitting of the majority class and efficiently learning high-frequency features.
    \item Extensive experiments on a real-world dataset show that SSLPDL can effectively capture spatiotemporal bias in forecasts, improving mean Intersection over Union (mIoU) by over 9.9\% compared to six popular benchmark models.
\end{itemize}

\begin{figure*}[t]
    \centering
    \includegraphics[width=0.95\textwidth]{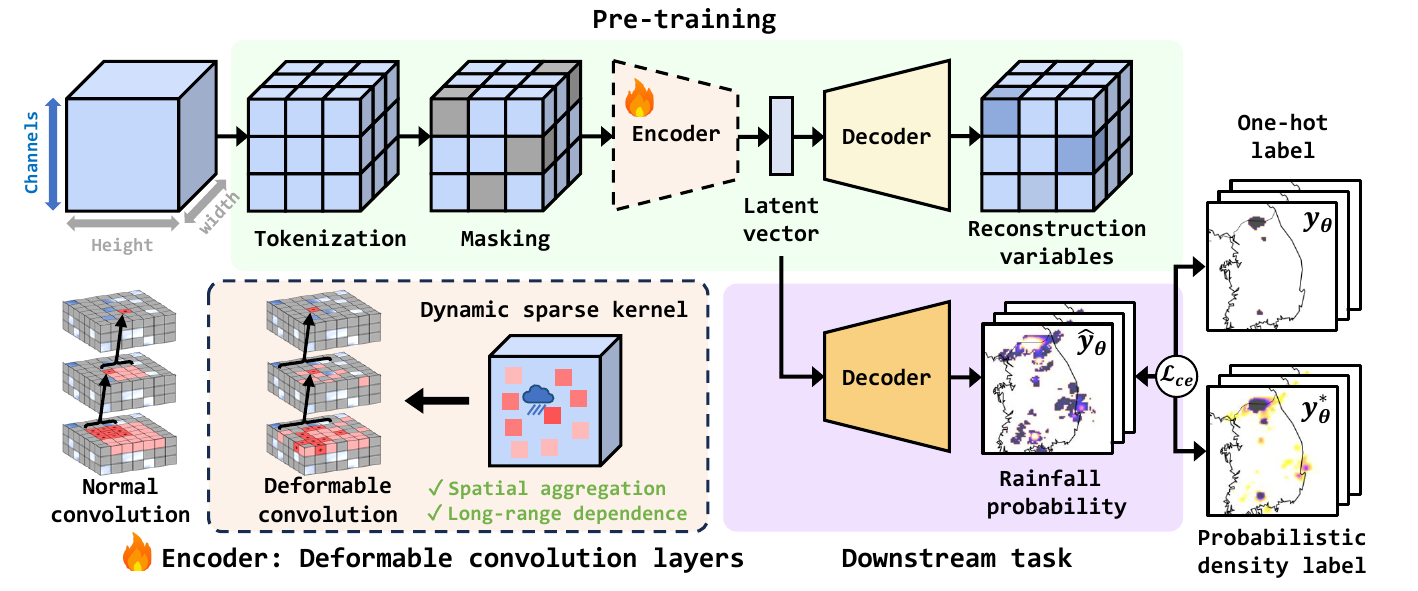}
    \caption{The overall structure of SSLPDL. Two-stage process representing spatiotemporal bias in forecasts: i) Pre-training and ii) Downstream task. Pre-training focuses on learning a variables-dependency by a reconstruction task. An encoder based on deformable convolution layers~\cite{wang:2023} is applied to capture and represent the spatial features from neighboring pixels effectively. Note that the deformable convolution aims to aggregate spatial features from the surrounding pixels to predict spatiotemporal bias. The downstream task utilizes the pre-trained encoder and probabilistic density labeling to estimate rainfall probability.}
    \label{fig:framework}
\end{figure*}

\section{Related Work}
\subsection{Masked Modeling in Atmospheric Science}
Masked modeling is a key area of research aimed at learning universal and shareable insights from vast datasets, thereby boosting performance in downstream tasks.
The masked auto-encoder (MAE)~\cite{he:2022} has emerged as a pivotal innovation in this field, significantly enhancing the ability to derive detailed feature representations from unlabeled data.
MAE introduces a refined pre-training approach for a vision Transformer (ViT) encoder~\cite{dosovitskiy:2020}, effectively addressing the long-range dependency challenges encountered in 3D video prediction tasks.
This approach enables effective visual representation learning through the simple pipeline of masking and reconstruction~\cite{tong:2022}. 
By reconstructing the masked patches, masked modeling facilitates an improved understanding of weather phenomena, providing a foundation for developing more accurate weather forecasting models~\cite{gong:2024}.

In global models, a pre-trained encoder from masked modeling can predict 3D atmospheric fields for medium-range time scales (typically 5 to 10 days).
These models can predict longer lead times because they do not need to account for updates to horizontal lateral boundary conditions.
However, precipitation forecasting remains a challenge~\cite{man:2023}.
Existing studies utilize 2D masked modeling to learn weather variables without adequately distinguishing the inherent characteristics of 3D datasets.
In addition, there is still limited research on pre-training schemes with masked modeling in atmospheric science. 
Our research builds upon these concepts by explicitly exploring the use of masked modeling for the representation learning of 3D variables, addressing long-range dependency problems.

\subsection{Class Imbalance}
Class imbalance refers to the issue where certain classes in a dataset have a much smaller number of samples compared to other classes~\cite{liu:2022}. 
This problem causes the model to be biased towards the majority class during the training process, degrading its generalization performance on the minority class~\cite{johnson:2019,chen:2023}. 
Precipitation datasets represent class imbalance, where heavy rain events occur much less commonly than no or light rain~\cite{aarthi:2022,tsagalidis:2022}. 
With climate change leading to increased storms and abnormal rainfall patterns~\cite{bria:2020,gao:2020}, class imbalance within the data has emerged as a significant issue in this research domain.
The problem of imbalanced learning has been addressed from three perspectives: loss reweighting, model ensembling, and data sampling~\cite{zhang:2024}. 
Loss reweighting methods, which learn weights specific to a given dataset, need help maintaining consistent performance across diverse precipitation datasets.
Model ensembling methods, on the other hand, are computationally expensive~\cite{wang:2020,zhang2:2021}. 
Consequently, this paper focuses primarily on data sampling. 
Prior studies have shown that data re-sampling can negatively impact representation learning by introducing redundant samples and increasing the risk of overfitting in the case of oversampling or by discarding valuable examples in the case of undersampling~\cite{mahajan:2018,kang:2019}.

Label smoothing is a regularization technique that can improve generalization and reduce overfitting~\cite{shwartz:2024}. 
This paper adapts data sampling with a labeling method to address the class imbalance problem.
Unlike conventional label smoothing method~\cite{muller:2019}, which typically assigns a small, fixed probability to all classes to prevent the model from becoming too confident in its predictions, our method ensures a more balanced representation of minority classes by dynamically adjusting probability values across different classes using probabilistic density labeling.
This approach leads to accurate probability distributions and reliable predictions of rare but critical weather events, such as severe storms or extreme weather phenomena.

\section{Self-Supervised Learning for Precipitation Post-processing}
In this section, we propose a self-supervised pre-training framework for precipitation post-processing.
The framework aims to learn the spatiotemporal biases in forecasts and improve the precipitation prediction performance.
Our framework utilizes deformable convolution layers to reconstruct input variables from masked inputs, facilitating the learning of spatiotemporal representations of precipitation bias.
We utilize the pre-trained encoder to post-process the precipitation dataset, increasing forecast accuracy and securing lead time.
Also, we introduce a probabilistic density labeling strategy to mitigate the class imbalance problem in predicting heavy rain.

\subsection{Problem Definition}
Given a multivariate forecast $X = \{x_1, \cdots, x_n\} \in \mathbb{R}^{n \times h \times w}$, we define the rainfall probability $Y = \{y_1, \cdots, y_c\} \in \mathbb{R}^{c \times h \times w}$, where $n$, $c$, $h$, and $w$ denote the number of variates, number of classes, height, and width, respectively.
Our objective is to derive a precipitation segmentation function $f_{\Theta}: \mathbb{R}^{n \times h \times w} \rightarrow \mathbb{R}^{c \times h \times w}$, involving reconstructing $f_{\phi}$ from the NWP forecasts to map the target object function $f_{\theta}$.
$\Theta$ is composed of $\phi$ and $\theta$, where $\phi$ refers to the reconstruction task $f_{\phi}: \mathbb{R}^{n \times h \times w} \rightarrow \mathbb{R}^{d \times h \times w}$ for the self-supervised pre-training, and $\theta$ represents the segmentation task $f_{\theta}: \mathbb{R}^{d \times h \times w} \rightarrow \mathbb{R}^{c \times h \times w}$ for predicting the rainfall probability, where $d$ denotes the dimensionality of the feature space after reconstruction.1

\subsection{Self-Supervised Pre-Training}\label{sec:pretraining}
Figure \ref{fig:framework} shows the overall framework of our self-supervised pre-training for precipitation post-processing.
The pre-training process for learning the variables-dependency entails the following three modules: (i) masking, (ii) encoder, and (iii) decoder.

\noindent\textbf{Masking.}
First, we tokenize the input variables $X \in \mathbb{R}^{n \times h \times w}$ into non-overlapping 3D patches $\mathcal{P} \in \mathbb{R}^{q \times p \times p}$, where $q$ and $p$ represent the number of channels and the size of each patch, respectively~\cite{fan:2021}.
These tokenized samples, denoted as $E$, are reshaped to $\mathbb{R}^{\frac{n}{q} \times \frac{h}{p} \times \frac{w}{p}}$.
We then use a 3D convolution layer to flatten and embed these tokens into a higher-dimensional space $\mathbb{R}^{N \times d}$, where $d=qp^{2}$ and $N=\frac{nhw}{d}$.
We form the masked input dataset $E_{mask}$ by combining the unmasked patches represented as $S$ with the masked patches ${\{\bar{S_j}}\}_{j=1}^{M}$, where each masked patch replaces its corresponding unmasked patch in the tokenized samples $E$.
The masked dataset, $E_{mask}$, combines these selectively obscuring input parts.
Finally, a learnable positional embedding $pe \in \mathbb{R}^{N \times d}$ is added to the tokenized samples to preserve positional information.
The masked patches are then restored to their original input dimensions of $E_{mask} \rightarrow X_{mask} \in \mathbb{R}^{n \times h \times w}$.
Our masking strategy employs a structure-agnostic approach~\cite{feichtenhofer:2022}, randomly applying masks with proportions ranging from 0 to 1.
This method often surpasses traditional structure-aware masking, focusing on specific spatial or channel features.

\noindent\textbf{Encoder.}
Our encoder's core is based on the InterImage, as described by Wang~\etal~\cite{wang:2023}.
This design uses deformable convolution layers incorporating sampling offsets to enable adaptive spatial aggregation and reduce over-inductive biases in the network.
Unlike prior methods that rely on fixed pixels, deformable convolutions effectively handle variable distribution shifts caused by accumulated errors inherent in the forecasts~\cite{tong:2022,gong:2024}.
Also, the hierarchical structure of InterImage is robust for long-range dependency, and it can achieve comparable or even better performance than vanilla Transformer \cite{vaswani:2017}.
Considering these advantages, we utilize InternImage as a backbone, aggregating the neighborhood of variables to enhance representation learning of accumulated errors.
The encoder includes layer normalization, feed-forward networks, and GELU activation.
It starts with feature extraction through a \textit{stem} network, followed by four stages of deformable convolution and down-sampling.
Each stage of our encoder maps to a hierarchy of hidden states $Z_{l\in (1, 4)}$ from the $X_{mask}$ with a stacking pattern of \textit{AABA}, which effectively learns richer information from different representation sub-spaces.

\noindent\textbf{Decoder.}
Our decoder aims to reconstruct the input by predicting the pixel values for each masked patch.
We employed the commonly used UperNet~\cite{xiao:2018} framework as a decoder to enable effective segmentation while preserving object boundaries and details from the training data.
The latent representation derived by hierarchical encoder layers $Z_{l}$ is mapped back to the multivariate input samples as $X \in \mathbb{R}^{n\times h \times w}$.
During training, the model predicts mask $\hat{X}$ and aims to minimize the difference between the predicted mask $\hat{X}$ and the input $X$ as follows:
\begin{equation}
    \mathcal{L}_{\text{rec}} = \frac{1}{|M|} \sum_{j \in M} \left\| \hat{X}_{j} - X_{j} \right\|_2^2,
\end{equation}
where the set $M$ indexes the masked in input $X$ for reconstruction, $\{\hat{X}_{j}\}_{j=1}^{M}$ and $\{X_{j}\}_{j=1}^{M}$ denote the reconstructed and original samples at the $j$-th masked patch, respectively.

\subsection{Precipitation Feature Representation}\label{sec:finetuning}
In this section, we introduce transfer learning using a pre-trained encoder as a feature extractor.
We employ UperNet as the decoder for semantic segmentation to predict rainfall probability.
The cross-entropy loss $\mathcal{L}_{ce}(\Theta; X)$ is computed as follow:
\begin{equation}
    \resizebox{.89\linewidth}{!}{$
        \mathcal{L}_{ce} = -\sum_{i=1}^{c} w_i \left(\beta y_{i} + (1-\beta) y^*_{i}\right) \log \left(\dfrac{\exp(\hat{y}_i)}{\sum_{j=1}^{c}\exp(\hat{y}_j)}\right),
        \label{eq:object_function}
    $}
\end{equation}
where $i$ denotes the index of the class label, and $j$ denotes the index of the predicted class. 
$y_i$ corresponds to the one-hot encoded true labels, $y^*_{i}$ represents the probabilistic density labels, and $\hat{y}_j$ denotes model's predictions. The weights $w_i$ are assigned to each class label to manage class imbalance, and $\beta$ is a hyperparameter that adjusts the trade-off between the fidelity to the one-hot labels and the distribution captured by the probabilistic density labels.
Minimizing the objective function leads to obtaining the maximum likelihood estimate of the segmentation model parameters.
This representation process yields parameters optimally fitting the data by minimizing the discrepancy between the model's predictions and the true labels.

\subsection{Probabilistic Density Labeling}
\label{sec:labeling}
The precipitation dataset represents the amount of liquid water that falls over a specific period.
These values can vary widely over arbitrary time intervals and are measured continuously.
Nonetheless, it is common practice to discretize precipitation datasets using thresholds for rainfall probability segmentation. 
One-hot labeling distinctly assigns a probability of 1 to a single class and 0 to all other classes, disregarding the inherent variability of rainfall intensity. 
This can lead to model instability and overfitting.
To tackle the problem, we present a straightforward approach to labeling precipitation datasets, as shown in Algorithm~\ref{alg1}, called probabilistic density labeling.
Our proposed probabilistic density labeling dynamically adjusts the probability values across different classes based on distributional assignment principles, adeptly capturing the subtle variations in rainfall intensity.

In meteorology, rainfall intensity is determined within predefined thresholds, which categorize precipitation amounts.
We fine-tune probability values based on proximity to these thresholds.
When rainfall $\gamma$ approaches the threshold $\tau$, the algorithm applies label smoothing to its probability distribution. 
This shift from rigid one-hot labeling to more continuous gradient-based adjustment.
The transformed function, denoted as $f_{prob}(y)$, converts the discrete one-hot label $y$ into the probabilistic density label $y^{*}$, is defined as:
\begin{equation}
\resizebox{0.9\hsize}{!}{$
    f_{prob}(y) = \underbrace{(1-\alpha)}_{\text{smoothing strength}} \cdot \underbrace{\dfrac{\tau_{i}-\gamma}{\tau_{i} - \tau_{i-1}}}_{\text{probability density}} + \underbrace{\dfrac{\alpha}{N}}_{\text{uniformity factor}},
$}
\label{eq:smoothing}
\end{equation}
where $\alpha$ represents a hyperparameter controlling the strength of label smoothing, $N$ denotes the number of classes, and $\tau_{i-1} \leq \gamma < \tau_{i}$ for $1 \leq i \leq N-1$. 
Note that the probability sum for each class equals 1.
Our labeling method ensures that probability values change gradually as they move from one threshold to another.

\begin{algorithm}[t]
\small
    \DontPrintSemicolon
    \caption{One-hot label and probabilistic density label}\label{alg1}
    \KwIn{Rainfall $\gamma \in [0, 100)$;\qquad\qquad\qquad\qquad Threshold set $\tau = [\tau_0, \ldots, \tau_{k-1}]$ \qquad\qquad\qquad\qquad\qquad where $0 < \tau_0 < \ldots < \tau_{k-1} < 100$\qquad\qquad\qquad\qquad (with $k$ as the number of thresholds)}
    \KwOut{One-hot label $y$; Probability density label $y^{*}$}
    \vspace{0.2cm}
    \tcp{Initialization}
    Set $N = k + 1$  \\
    Set smoothing parameter $\alpha$ \\
    Initialize $y, y^{*} = [0, \cdots, 0]$ with length $N$ \\
    
    \If{$N < 2$}{
        \Return{Error: Number of classes must be at least 2}
    }
    
    \vspace{0.2cm}
    
    \tcp{One-hot label $y$}
    \For{$i \gets 0$ \textbf{to} $N-1$}{
        \textbf{if} $i = 0$ \textbf{and} $\gamma < \tau_{0}$ \textbf{ then } $y_{0} \gets 1$\\
        \textbf{else if} $i = N-1$ \textbf{and} $\gamma \geq \tau_{k-1}$ \textbf{ then } $y_{N-1} \gets 1$\\
        \textbf{else if} $\tau_{i-1} \leq \gamma < \tau_{i}$ \textbf{ then } $\ y_{i} \gets 1$\\
        \textbf{else} $\ y_{i} \gets 0$
    }
    
    \vspace{0.2cm}
    
    \tcp{Probabilistic density label $y^{*}$}
    \For{$i \gets 0$ \textbf{to} $N-1$}{
        \textbf{if} $0 \leq \gamma < \tau_0$ \textbf{ then } 
            $y^{*}_{0} \gets (1 - \alpha) \dfrac{\tau_0 - \gamma}{\tau_0} + \dfrac{\alpha}{N}$ \\
        \textbf{else if} $\tau_{i-1} \leq \gamma < \tau_i$ \textbf{ then } 
                $y^{*}_{i} \gets (1 - \alpha) \left(1 - \dfrac{\gamma - \tau_{i-1}}{\tau_i - \tau_{i-1}}\right) + \dfrac{\alpha}{N}$ \\
        \textbf{else if} $\tau_{k-1} \leq \gamma < 100$ \textbf{ then } 
                $y^{*}_{N-1} \gets (1 - \alpha) + \dfrac{\alpha}{N}$ \\
        \textbf{else}
            $y^{*}_{i} \gets \dfrac{\alpha}{N}$ \\
    }
    \Return{$y, y^{*}$}
\end{algorithm}

\section{Experiments}
\noindent{\textbf{Baselines.}} We compare our SSLPDL on the regional precipitation dataset, called RDAPS, with six models over forecasting: \textbf{RDAPS (NWP)}, \textbf{ConvLSTM}~\cite{shi:2015}, \textbf{Metnet}~\cite{sonderby:2020}, \textbf{MAE}~\cite{feichtenhofer:2022}, \textbf{Swin-Unet}~\cite{cao2022swin}, and  \textbf{PostRainBench}~\cite{tang2023postrainbench}\footnote{\small{Appendix A provides a more detailed description of the experiment setting.}}. 
Experimental results show remarkable accuracy in predicting rainfall probability and excel in forecasting rare events with above 10 mm rainfall.
Furthermore, the proposed methods secure an extended lead time while maintaining forecast precision. 

\subsection{Dataset}
We utilized the Korean regional data assimilation and prediction system (RDAPS) for the input weather variables to extract forecasts spanning 25 to 30 hours, targeting a 24-hour prediction.
The RDAPS forecasts consist of 3 km resolution data covering East Asia.
The variables are curated under the guidance of previous studies~\cite{pathak:2022,man:2023,keisler:2022}, as shown in Table \ref{tab:data_table}.
All variables were adopted for $z$-score normalization for each variable fragment using global statistics of a sampled set.
Additionally, separate normalization applies for variables with the same attributes but different vertical levels.
We used quantitative precipitation estimation (QPE) as ground truth, which is output data that estimates the actual value of precipitation measured by remote sensors (radar and satellite) and ground observers (automatic weather stations) based on various algorithms.
The precipitation reanalysis site is one of the global standards for monitoring weather phenomena. 
In Korea, data is provided with a 5 km resolution every hour.
The ground truth covers the terrain and surrounding sea of the Korean peninsula with the size of 253 $\times$ 149.

\begin{table}[t]
\centering
\small
\renewcommand{\arraystretch}{0.9}
\begin{tabular}{P{2.5cm}P{5cm}}
     \toprule
     Vertical level & Variables \\
     \midrule
     Surface & $U_{10}, V_{10}, T_{2}, SLP, rain$ \\
     850 $hPa$ & $U, V, T, Z, RH$  \\
     500 $hPa$ & $U, V, T, Z, RH$\\
     100 $hPa$ & $Z$\\
    \bottomrule
\end{tabular}
\caption{Input data summarization. Our input data comprises 16 variables, including seven physical variables across four pressure levels: temperature (T), U-component of the wind speed (U), V-component of the wind speed (V), geopotential height (Z), relative humidity (RH), sea-level pressure (SLP), and rain.}
\label{tab:data_table}
\end{table}

\noindent\textbf{Data sampling.}
In this work, the training dataset spans from June 2020 to May 2022, and the validation dataset spans from September to November 2022.
For the test dataset, we collected August 2022 in South Korea. 
Given Korea's high annual precipitation levels, with more than 50\% occurring during August, this dataset is particularly valuable for studying rainfall patterns.
We pre-trained the model on all dataset samples, excluding data from the winter season and the validation and test datasets.
We remapped the NWP forecasts into QPE grids using the nearest neighbor method and resized the input data to match the dimensions of the QPE.
For consisting of the non-overlapped patches, we conducted a center crop of the dataset into the size of 224 $\times$ 128.

\noindent\textbf{Data categorization.}
We categorize rainfall into three groups based on studies on the standard criterion~\cite{wmo:2008}, utilizing thresholds of 0.1 mm and 10 mm.
Table \ref{tab:proportion} illustrates the proportions of each class. 
Within the training dataset, the proportion of heavy rain (group 3) was approximately 0.5\%, highlighting a notable class imbalance. 
This prevalent issue in precipitation measurement data presents a common challenge in forecasting. 
To mitigate this issue, we initially addressed the problem through data sampling. 
We conducted data sampling with rainy days comprising 80\% and non-rainy days 20\% of the dataset. 
Subsequently, we employed probabilistic density labeling to address the class imbalance problem.

\begin{table}[t]
    \centering
    \small
    \renewcommand{\arraystretch}{0.9}
    \begin{tabular}{lcccc}
         \toprule
         Notes & Rainfall & Dataset &Sampling & Labeling \\
         \midrule
         Group 1 & (0, 0.1] & 89.89 & 83.97 & 73.58\\
         Group 2 & (0.1, 10] & 9.61 & 15.22 & 13.33\\
         Group 3 & (10, $\infty$) & 0.50 & 0.81 & 13.09\\
        \bottomrule
    \end{tabular}
    \caption{Summary of label proportions in the training dataset. \textbf{Notes}: Groups of class (group 1: no rain, group 2: rain, group 3: heavy rain), \textbf{Rainfall}: 1-hour cumulative rainfall [mm], \textbf{Dataset}: Original data proportion, \textbf{Sampling}: Data proportion after data sampling, \textbf{Labeling}: Data proportion after probabilistic density labeling proposed in Section \ref{sec:labeling}.}
    \label{tab:proportion}
\end{table}

\begin{table*}[ht]
    \centering
    \small
    \resizebox{\textwidth}{!}{
    \begin{tabular}{P{2.6cm}|P{1.4cm}|*{4}{P{1.1cm}}|*{4}{P{1.1cm}}|P{1.1cm}|P{1.4cm}}
        \hline
        \multirow{2}{*}{Method} & \multirow{2}{*}{Post-proc.} & \multicolumn{4}{c|}{CSI$_{0.1}$ $\uparrow$ (rain)} & \multicolumn{4}{c|}{CSI$_{10}$ $\uparrow$ (heavy rain)} & \multirow{2}{*}{mIoU $\uparrow$} & \multirow{2}{*}{\#Param. (M)} \\ 
        \cline{3-10}    
        &  & CSI & F1 & Precision & Recall & CSI  & F1 & Precision & Recall & \\
        \hline
        \raggedright RDAPS & \xmarkcol & 0.295 & 0.408 & 0.364 & \textbf{0.465} & \underline{0.060} & 0.113 & 0.096 & 0.136 & 0.380 & -\\
        \raggedright ConvLSTM*~\cite{shi:2015} & \cmarkcol & 0.285 & 0.358 & 0.412 & 0.358 & 0.001 & 0.002 & 0.001 & 0.129 & 0.343 & 6.3\\
        \raggedright Metnet*~\cite{sonderby:2020} & \cmarkcol & 0.323 & 0.456 & 0.500 & 0.420 & 0.007 & 0.015 & 0.007 & \textbf{0.189} & 0.367 & 2.5\\
        \raggedright MAE*~\cite{feichtenhofer:2022} & \xmarkcol & 0.338 & 0.435 & 0.432 & 0.438 & \underline{0.060} & \underline{0.118} & \underline{0.206} & 0.077 & 0.384 & 329\\
        \raggedright Swin-Unet*~\cite{cao2022swin} & \xmarkcol & 0.356 & 0.484 & \underline{0.553} & 0.431 & 0.042 & 0.081 & 0.061 & 0.122 & \underline{0.389} & 27.2\\
        \raggedright PostRainBench*~\cite{tang2023postrainbench} & \cmarkcol & \underline{0.358} & \underline{0.486} & 0.479 & \underline{0.457} & 0.028 & 0.055 & 0.069 & 0.098 & \underline{0.389} & 27.2\\
        \rowcolor[HTML]{E0E0E0} \raggedright \textbf{Ours*} & \cmarkcol & \textbf{0.392} & \textbf{0.497} & \textbf{0.575} & 0.439 & \textbf{0.096}  & \textbf{0.174} &  \textbf{0.219} & \underline{0.145} & \textbf{0.412} & 49.5\\
        \hline
    \end{tabular}}
    \caption{Quantitative comparison on RDAPS test set, using CSI and mIoU. The results were calculated as averages over forecast times ranging from 25 to 30 hours. * denotes that probabilistic density labeling is applied, and Post-proc. denotes \textit{post-processing approach}. \textit{Post-processing approach} represents the official code aligned with the model structures for the post-processing task.}
    \label{tab:result}
\end{table*}

\subsection{Evaluation Metrics} 
We utilize four evaluation metrics: the critical success index (CSI), F1 score, Precision, and Recall.
CSI is commonly used in precipitation forecasting to evaluate the accuracy of forecasts, explicitly focusing on the successful prediction of events~\cite{shi:2015,sonderby:2020,tang2023postrainbench}. 
This metric assesses the agreement between observed and predicted events and strategically highlights the severe implications of failing to predict significant occurrences. The CSI is defined as follows:
\begin{equation}
    \text{CSI} = \dfrac{\text{TP}_{\tau\scalebox{0.6}{$\leq$}}}{\text{TP}_{\tau\scalebox{0.6}{$\leq$}}+\text{FP}_{\tau\scalebox{0.6}{$\leq$}}+\text{FN}_{\tau\scalebox{0.6}{$\leq$}}},
\end{equation}
\noindent where $\text{TP}_{\tau\scalebox{0.6}{$\leq$}}$, $\text{FP}_{\tau\scalebox{0.6}{$\leq$}}$, and $\text{FN}_{\tau\scalebox{0.6}{$\leq$}}$ represent the number of true positives, false positives, and false negatives, respectively, relative to the predefined threshold $\tau$.
Additionally, we utilized the mIoU to analyze segmentation tasks, quantifying the overlap between predicted and ground truth areas.
mIoU calculates the ratio of correct forecasts to the total number of occasions. 
We avoid using \textit{Accuracy} due to the class imbalance issue, where heavy precipitation exceeding 10 mm is rare.
In such cases, a model that predicts no values exceeding the threshold may outperform another model.

\begin{figure}[t]
    \centering
    \includegraphics[width=\columnwidth]{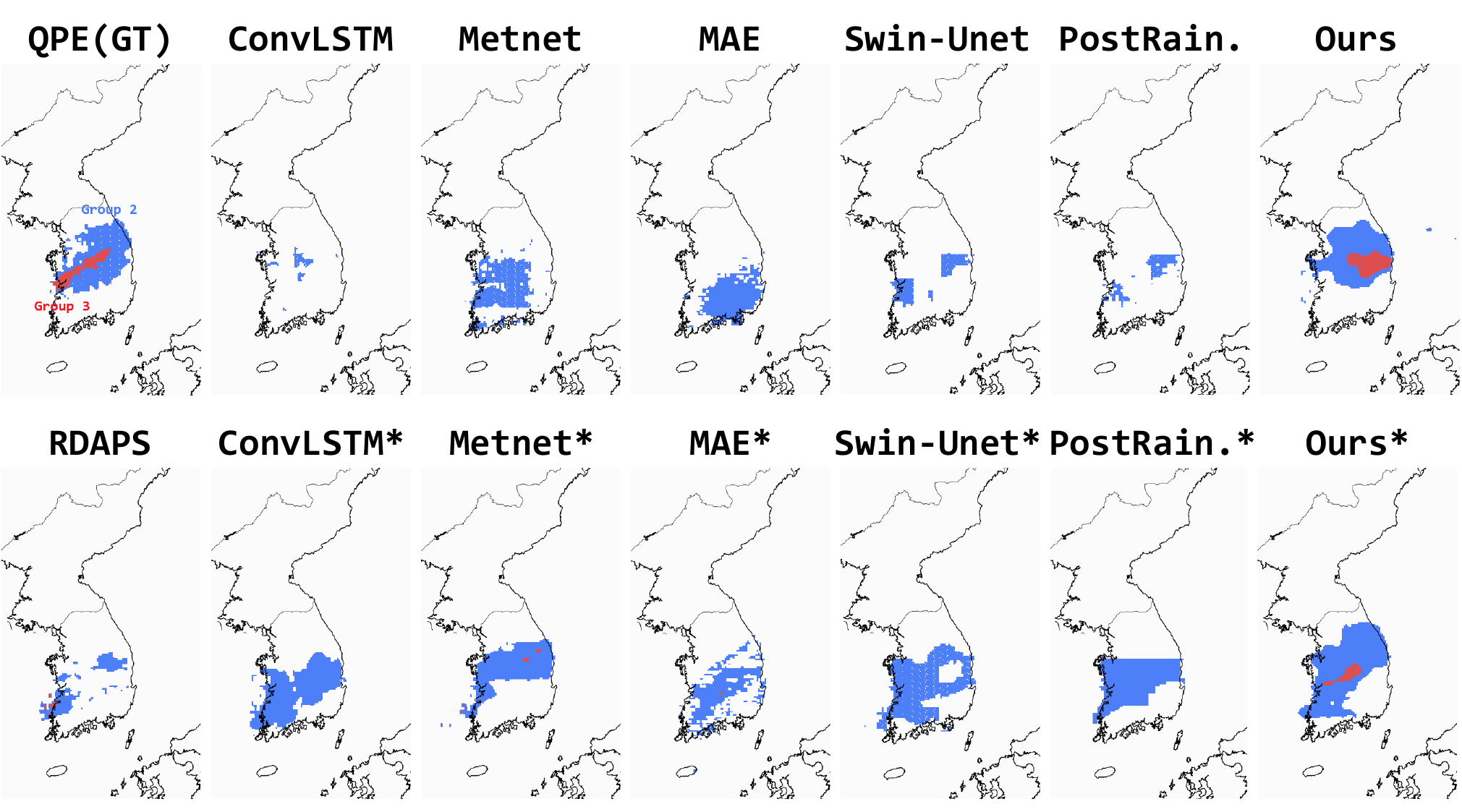}
    \caption{Visualization result between benchmarks on August 15 2022 at 18 UTC (+29 h). Colors represent each group (group 1: white, group 2: blue, and group 3: red). As the stationary front moved southward, cold air from the northwest entered the upper atmosphere while recent rainfall left abundant moisture near the surface, raising temperatures. This caused atmospheric instability, resulting in sporadic showers typical of localized precipitation events. Despite regional variations in rainfall intensity and amount, the proposed method accurately predicts concentrated heavy rain. * denotes that probabilistic density labeling is applied.}
    \label{fig:visual}
\end{figure}

\subsection{Main Results}
As shown in Table~\ref{tab:result}, our model outperforms all the baselines on the RDAPS test set.
Across all rainfall intervals, the CSI is higher for our model than all the baselines.
We confirm that our model can accurately detect rain events while minimizing the number of false positives and missed rain events.
Our model has reliable accuracy across the board at above 0.1 mm rainfall compared to other models. 
Above 10 mm rainfall, our model consistently performs well, suggesting its reliability in predicting heavy rain events.
Additionally, compared with the baselines, augmenting spatial aggregation has been determined to facilitate the learning of significant spatial correlation within spatiotemporal data.
We have observed that the proposed model effectively adapts to various patterns and scales in challenges such as precise heavy rain, showcasing its versatility and efficacy in managing complex spatiotemporal scales and patterns.
We addressed the class imbalance issue by implementing probabilistic density labeling. 
Table~\ref{tab:result} and Table~\ref{tab:labeling} present that the proposed labeling method enhances performance and mitigates the imbalance problem across benchmarks.

Figure \ref{fig:visual} shows that above 0.1 mm of precipitation covers much of the Korean Peninsula. 
Our model captures above 10 mm rainfall, while NWP models (RDAPS) underestimate this case.
Also, we observed that our model well predicts rainfall location and above 10 mm rainfall, but these tend to be overestimated.
The results show that combining $y$ and $y^{*}$ will further enhance the accuracy.
The limitation is that the input data relies on an NWP precipitation variable. 
As a result, the prediction model learns with a certain margin of error and predicts like RDAPS in some cases.

\begin{table}[t]
    \centering
    \small
    \begin{tabular}{*{3}{P{0.95cm}}|*{3}{P{0.95cm}}}
    \hline
    \multicolumn{3}{c|}{Pre-training} & \multicolumn{3}{c}{Transfer learning}\\
    \hline
    \small{Ratio} & \scriptsize{CSI$_{0.1}$ $\uparrow$} & \scriptsize{CSI$_{10}$ $\uparrow$} & \small{Ratio} & \scriptsize{CSI$_{0.1}$ $\uparrow$} & \scriptsize{CSI$_{10}$ $\uparrow$}\\
    \hline
    50\% & 0.310 &  0.088 & 0\% & 0.389 & 0.086\\
    75\% & \textbf{0.389} & 0.086 & 25\% & \textbf{0.392} & \textbf{0.096}\\
    90\% & 0.356 &  \textbf{0.089} & 50\% & 0.339 & 0.054\\
    \hline
    \end{tabular}
    \caption{Ablation study on the masking ratio in pre-training and transfer learning. CSI$_{(\cdot)}$ represents the CSI score with each threshold [mm].}
    \label{tab:pre_mask}
\end{table}

\subsection{Ablation Studies}
\noindent\textbf{Masking ratio.}
We compared the masking ratio results during pre-training and transfer learning to determine the optimal masking ratio for precipitation post-processing. 
Table~\ref{tab:pre_mask} shows that the larger the masking ratio in the pre-training, the better the results. 
When the masking ratio was increased to 90\%, each feature exhibited smoother restoration results, likely influencing the segmentation task. Further details on the reconstruction results are provided in Appendix C.
During the fine-tuning process, observing a smaller masking ratio of 25\% in transfer learning led to better results.
Given that our dataset incorporates errors from time-series forecasting, we posit that masking aids the model in prioritizing relevant variable dependencies while disregarding irrelevant or temporal errors. 
Moreover, masked modeling serves as a form of regularization, preventing the model from memorizing noise or outliers in the input data, thereby enhancing performance on test data.

\noindent\textbf{Pre-training analysis.}
This section investigates how a pre-training process that learns variable dependencies affects performance.
We compared our model with the scratch model (no pre-training process).
Table \ref{tab:labeling} compares model performance when employing pre-trained models with the reconstruction task.
Our model improves performance on segmentation tasks with the pre-training process, surpassing the same model without pre-training on the entire training set.
Remarkably, not only does the pre-trained model outperform the scratch model in heavy rain, but the benefit persists throughout the mIoU for the entire class.

\begin{table}[t]
\centering
    \small
    \resizebox{\columnwidth}{!}{
    \begin{tabular}{P{1.8cm}|*{3}{P{1.62cm}}}
        \hline
        Method &  CSI$_{0.1}$ $\uparrow$ & CSI$_{10}$ $\uparrow$ & mIoU $\uparrow$  \\
        \hline
        \raggedright ConvLSTM & .095 (\tri.190) & .000 (\tri.001) & .307 (\tri.036)\\
        \raggedright Metnet & .271 (\tri.052) & .006 (\tri.001) & .320 (\tri.047) \\
        \raggedright MAE & .312 (\tri.025) & .038 (\tri.022) & .373 (\tri.011)\\
        \raggedright Swin-Unet & .295 (\tri.061) & .006 (\tri.036) & .373 (\tri.016)\\
        \raggedright PostRainBench & .274 (\tri.084) & \textbf{.063 (\trid.035)} & .375 (\tri.014)\\
        \hline
        \raggedright Scratch & .338 (\tri.010) & .027 (\tri.040) & .381 (\tri.006)\\
        \rowcolor[HTML]{E0E0E0} \raggedright \textbf{Ours} & \textbf{.381 (\tri.011)} & .042 (\tri.054) & \textbf{.396 (\tri.016)}\\
        \hline
    \end{tabular}}
    \caption{Ablation study on probabilistic density labeling in RDAPS test set. ($\cdot$) denotes improved scores by probabilistic density labeling. \textit{Scratch} represents no pre-training, and $\beta$ was set to 0.25.}
    \label{tab:labeling}
\end{table}

\begin{figure}[b]
    \centering
    \begin{subfigure}[b]{0.45\columnwidth} 
        \caption{Optimization of $\alpha$}
        \includegraphics[width=\linewidth]{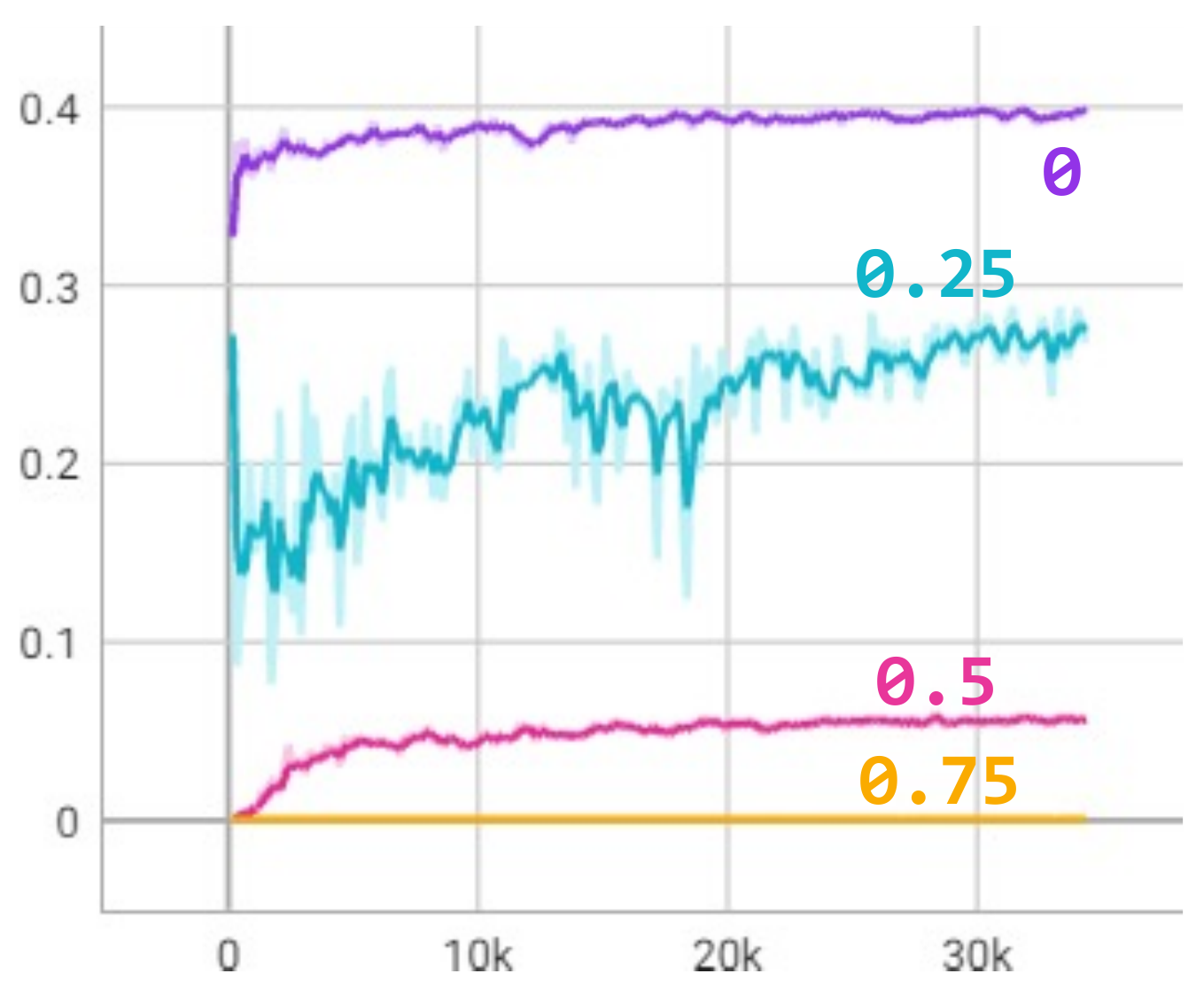}
        \label{fig:gradient-alpha}
    \end{subfigure}
    \hfill 
    \begin{subfigure}[b]{0.45\columnwidth}
        \caption{Optimization of $\beta$}
        \includegraphics[width=\linewidth]{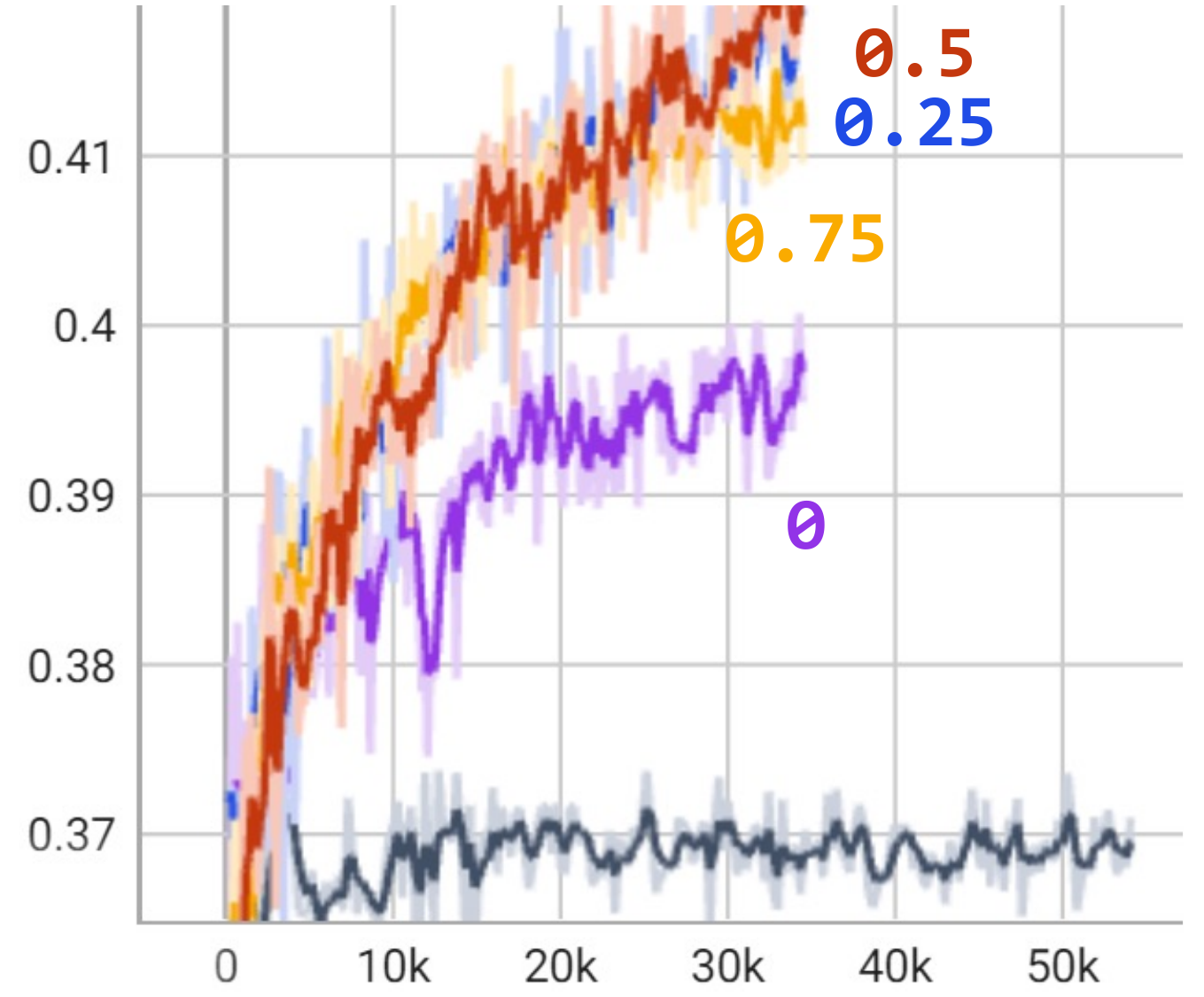}
        \label{fig:gradient-beta}
    \end{subfigure}
    \vspace{-5mm}
    \caption{Ablation study on hyperparamter for optimizing probabilistic density labeling. The y-axis represents the mIoU. (a) Gradient $\alpha$ of label smoothing. (b) Ratio $\beta$ of loss function between one-hot labels $y$ and probabilistic density labels $y^{*}$.}
    \label{fig:gradient}
\end{figure}

\noindent\textbf{Probabilistic density labeling analysis.}
In Figure \ref{fig:gradient-alpha}, we observe that smoothing significantly limits training and generalization.
Compared with the no smoothing ($\alpha$=0) parameter in Equation \ref{eq:smoothing}, which strongly focuses on heavy rain, other smoothed results show lower performance. 
Notably, smoothing reduces confidence, even for pixels without rain, due to performance degradation hindering their prediction ability.
Figure \ref{fig:gradient-beta} illustrates that employing one-hot labeling alongside probabilistic density labeling leads to improved model performance, with the density labeling alone yielding superior results.
Our analysis of the results, particularly regarding the $\beta$ ratio, reveals that an excessively high ratio of probabilistic density labeling tends to be learned as true positives for many areas, resulting in an overemphasis on heavy rain.
This issue can be alleviated by integrating losses from the one-hot labeling with those from the probabilistic density labeling, allowing for a more elaborated representation through the accurate labeling of true negatives.

\begin{table}[t]
\small
\centering
    \resizebox{\columnwidth}{!}{
    \begin{tabular}{P{0.2cm}|P{1.05cm}|*{5}{P{0.8cm}}}
        \hline
        & \multirow{2}{*}{Labeling} & Control & \multicolumn{2}{c}{Undersamp.} &  \multicolumn{2}{c}{Oversamp.}\\
        & & 90\% & 80\% & 70\% & 80\% & 70\% \\
        \hline
        \multirow{3}{*}{\rotatebox{90}{mIoU}} & - & 0.396 & 0.392 & 0.393 & 0.390 & 0.403 \\
        & Smooth. & 0.400 & 0.392 & 0.399 & 0.399 & 0.408 \\
        & \cellcolor[HTML]{E0E0E0} \textbf{Ours} & \cellcolor[HTML]{E0E0E0} \textbf{0.412} & \cellcolor[HTML]{E0E0E0} \textbf{0.402} & \cellcolor[HTML]{E0E0E0} \textbf{0.397} & \cellcolor[HTML]{E0E0E0} \textbf{0.403} & \cellcolor[HTML]{E0E0E0} \textbf{0.417}\\
        \hline
        \multirow{3}{*}{\rotatebox{90}{CSI$_{0.1}$}} & - & 0.381 & 0.366 & 0.384 & 0.370 & 0.386 \\
        & Smooth. & 0.364 & 0.379 & 0.373 & \textbf{0.382} &  0.391 \\
        & \cellcolor[HTML]{E0E0E0} \textbf{Ours} & \cellcolor[HTML]{E0E0E0} \textbf{0.392} & \cellcolor[HTML]{E0E0E0} \textbf{0.381} & \cellcolor[HTML]{E0E0E0} \textbf{0.375} & \cellcolor[HTML]{E0E0E0} 0.381 & \cellcolor[HTML]{E0E0E0} \textbf{0.396}\\
        \hline
        \multirow{3}{*}{\rotatebox{90}{CSI$_{10}$}} & - & 0.042 & 0.060 & 0.074 & 0.050 & 0.061 \\
        & Smooth. & 0.066 & 0.060 & 0.078 & 0.062 & 0.080 \\
        & \cellcolor[HTML]{E0E0E0} \textbf{Ours} & \cellcolor[HTML]{E0E0E0} \textbf{0.096} & \cellcolor[HTML]{E0E0E0} \textbf{0.096} & \cellcolor[HTML]{E0E0E0} \textbf{0.089} & \cellcolor[HTML]{E0E0E0} \textbf{0.078} & \cellcolor[HTML]{E0E0E0} \textbf{0.103}\\
        \hline
    \end{tabular}}
    \caption{Ablation study on sampling methods in RDAPS test set. `Smooth.' refers to traditional label smoothing \cite{muller:2019}. Control: Maintains an 8:2 ratio of no rain to rain, with approximately a 1:9 pixel ratio. Undersamp.: Under-sampling with no rain pixels at approximately 80\% and 70\%. Upsamp.: Up-sampling with data augmentation to set no rain pixels at approximately 80\% and 70\%.}
    \label{tab:sampling}
\end{table}

\noindent\textbf{Sampling.}
We address the issues of overfitting and noise in minority class data by utilizing under-sampling and over-sampling approaches. By modifying the training data distribution through sampling, we aim to reduce imbalance and mitigate noise caused by outliers. Table \ref{tab:sampling} presents the performance comparison between the proposed probabilistic density labeling technique and the traditional label smoothing technique \cite{muller:2019} when applied to sampling.
We employed six data augmentation techniques to increase the samples with higher rainfall for over-sampling. 
These techniques include flipping, resizing \cite{mehari:2022}, mixup \cite{zhang:2017}, and Gaussian noise.
Our experimental results indicate that the proposed probabilistic density labeling technique generally outperforms label smoothing, especially in pixels with heavy rain exceeding 10 mm.
For 80\% under-sampling, it shows better generalization performance than the original results, whereas 70\% under-sampling results in performance degradation. 
On the other hand, over-sampling yields better performance as the proportion of rain pixels increases. 
This analysis suggests that under-sampling may reduce generalization performance due to the smaller dataset size.
Additionally, data augmentation involves enriching knowledge learning with additional information or details, improving the accuracy or reliability of the model in the weather data.

\begin{figure}[t]
    \centering
    \includegraphics[width=\columnwidth]{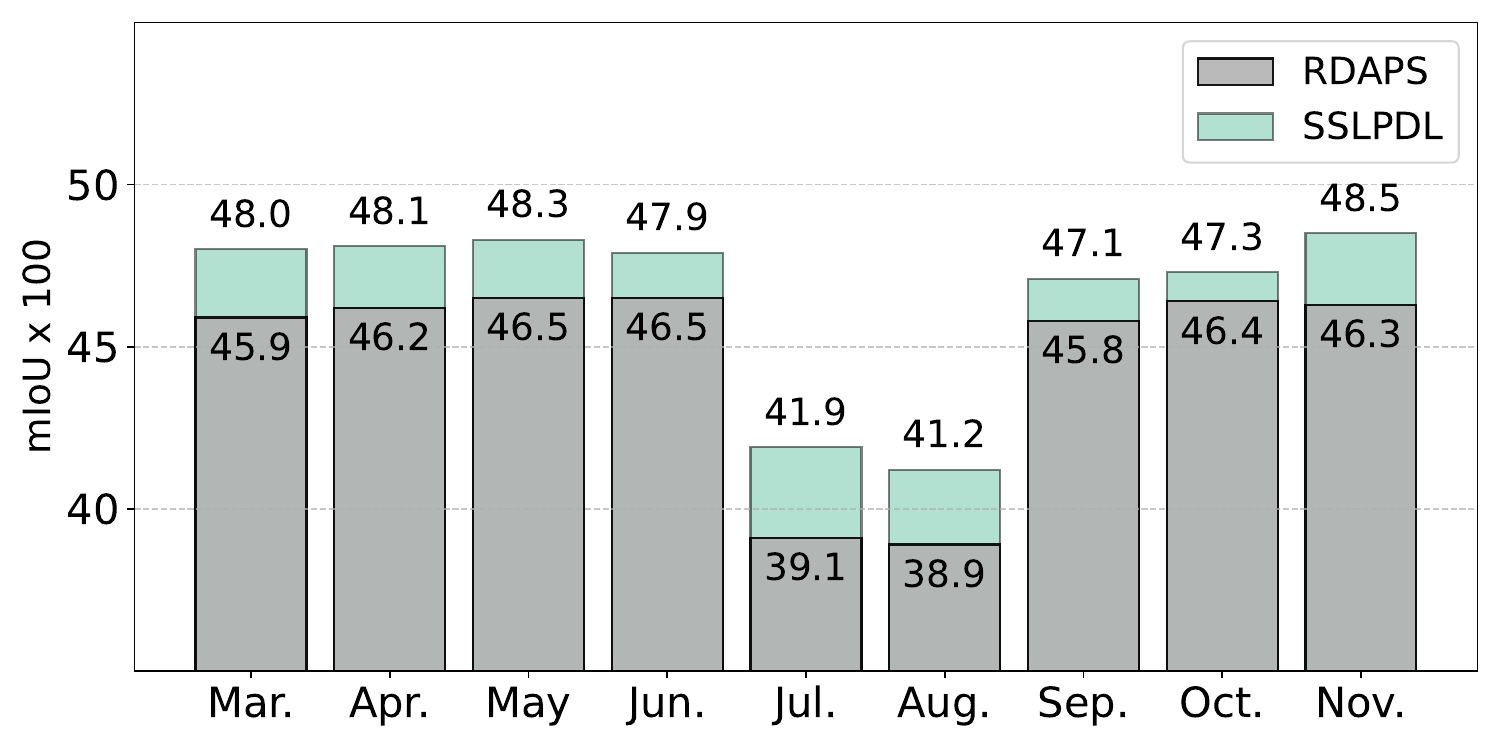}
    \caption{Monthly analysis of the impact of our SSLPDL on RDAPS.}
    \label{fig:applicapability}
\end{figure}

\noindent\textbf{Analysis of monthly impact.} 
In the Republic of Korea, approximately 66\% of annual precipitation and 90\% of heavy rainfall occur from June to September, driven by the influence of the North Pacific high. 
During this monsoon period, the interaction of five air masses generates diverse precipitation patterns, making it a critical season for understanding summer rainfall variability in East Asia. 
As part of this study, we conducted experiments for all seasons except winter, as precipitation data is unavailable. 
Notably, the summer months of July and August fall within the monsoon season known as ``Changma.'' 
This period is characterized by high variability and unpredictability, which makes long-term precipitation forecasting particularly challenging \cite{an2024self, lee2019validation}.
Figure~\ref{fig:applicapability} shows the capability of SSLPDL to achieve an overall performance improvement of approximately 4.2\% over RDAPS, with a notable gain of 7.1\% during July, a month characterized by high variability and uncertainty.
These results indicate the potential practical applicability of SSLPDL in regions heavily influenced by monsoons. 

\section{Conclusion}
In this work, we present a self-supervised pre-training approach for rainfall probability estimation by post-processing NWP forecasts, utilizing masked modeling to reconstruct atmospheric variables. 
We propose a method called SSLPDL that uses deformable convolution to finely account for the distribution shifts in each variable caused by accumulated errors in NWP models while enabling adaptive spatial aggregation. 
Additionally, we introduce a straightforward labeling method based on probability density to address the inherent variability within rainfall intensities and the issue of class imbalance in the precipitation dataset. 
Experimental results on the regional precipitation dataset show that SSLPDL outperforms other benchmarks and demonstrates competitive performance in securing forecast lead time.

\section*{Acknowledgement}
This work was supported by the Technology Innovation Program (RS-2024-00445759, Development of Navigation Technology Utilizing Visual Information Based on Vision-Language Models for Understanding Dynamic Environments in Non-Learned Spaces) funded by the Ministry of Trade Industry \& Energy (MOTIE, Korea).
We gratefully acknowledge the Korea Institute of Atmospheric Prediction Systems for their provision of the data and experimental facilities essential to this research.

{\small
\bibliographystyle{ieee_fullname}
\bibliography{egbib}
}

\end{document}